\documentclass[10pt,twocolumn,letterpaper]{article}
\usepackage[T1]{fontenc}
\usepackage[utf8]{inputenc}
\usepackage[margin=0.75in,columnsep=0.25in]{geometry}
\usepackage{times}
\usepackage{amsmath,amssymb,graphicx,booktabs,longtable,array}
\usepackage[round,authoryear]{natbib}
\usepackage[hyphens]{url}
\usepackage[hidelinks]{hyperref}
\makeatletter
\newenvironment{appendixfigure}{\par\medskip\noindent\begin{minipage}{\linewidth}\centering\def\@captype{figure}}{\end{minipage}\par\medskip}
\newenvironment{appendixtable}{\par\medskip\noindent\begin{minipage}{\linewidth}\centering\def\@captype{table}}{\end{minipage}\par\medskip}
\makeatother
\hypersetup{
  pdftitle={View Matters: Keyframe-Guided Text-Driven 3D Gaussian Editing},
  pdfauthor={Kaizhe Zhang; Yijie Zhou; Weizhan Zhang; Xuanyu Wang; Feng Lei; Sha Gong},
  pdfsubject={Text-driven 3D Gaussian editing},
  pdfkeywords={3D Gaussian Splatting, text-driven editing, keyframes}
}
\title{\LARGE\bfseries View Matters: Keyframe-Guided\\Text-Driven 3D Gaussian Editing}
\author{%
\normalsize Kaizhe Zhang\textsuperscript{1}, Yijie Zhou,
Weizhan Zhang\textsuperscript{1}\thanks{Corresponding author.},
Xuanyu Wang, Feng Lei, Sha Gong\\[6pt]
\small \textsuperscript{1}School of Computer Science and Technology, MOE-KLINNS Lab\\
\small Xi'an Jiaotong University, Xi'an, Shaanxi 710049, China\\
\small \texttt{zkz1081@stu.xjtu.edu.cn}, \texttt{zhangwzh@xjtu.edu.cn}}

\date{}
\begin{document}
\maketitle
\thispagestyle{plain}

\begin{abstract}
Text-driven 3D Gaussian editing commonly does not distinguish the editing reliability of rendered views, although different viewpoints provide supervision of substantially different quality. Views that clearly show the scene and match the edit instruction provide reliable guidance, while less informative views may weaken the edit when all views are treated equally. We present \textbf{View Matters}, a view-importance-aware framework that conducts editing around reliable keyframes. Keyframe Importance Estimation (KIE) identifies reliable views using geometric visibility, semantic distinctiveness, and edit relevance. Keyframe-Guided Editing (KGE) then propagates their editing signals asymmetrically to non-keyframes without noisy reverse influence, while Importance-Aware Optimization (IAO) preserves this reliability preference during 3DGS optimization. Across 23 scene--prompt pairs, View Matters achieves the highest average CLIP text--image similarity of 0.2822 and directional similarity of 0.2564 among the evaluated methods, with a four-minute editing time. Additional adjacent-view analysis indicates that the fidelity-oriented editing process maintains cross-view coherence.
\end{abstract}

\section{Introduction}

Text-driven 3D editing modifies reconstructed scenes from natural-language instructions. Neural Radiance Fields (NeRFs) and 3D Gaussian Splatting (3DGS) enable this task through differentiable rendering and pretrained image editors~\cite{haque2023instruct,wang2024gaussianeditor}. In particular, the explicit representation and efficient rendering of 3DGS make it well suited for iterative 3D editing~\cite{kerbl20233d}.

Most text-driven 3DGS editing methods follow a render--edit--optimize pipeline: a scene is rendered from multiple viewpoints, the rendered images are edited by a pretrained 2D model, and the edited observations are used to optimize the 3D representation~\cite{wang2024gaussianeditor,wu2024gaussctrl}. A limitation of this paradigm is that editing supervision is not equally reliable across viewpoints. Informative views provide faithful editing signals, whereas ambiguous views may produce weak or inaccurate edits. Treating all views equally allows these unreliable signals to propagate into and accumulate during 3DGS optimization, reducing the fidelity of the edit.

Existing methods improve multi-view agreement through depth guidance, feature fusion, epipolar interaction, or shared attention~\cite{wu2024gaussctrl,chen2024dge,lee2025editsplat}. Some methods already designate reference or key views for cross-view propagation. However, they do not explicitly determine which views should dominate the editing process. Symmetric interaction or uniform supervision, therefore, allows less reliable views to influence more reliable ones, making the final edit less faithful to the instruction. Consequently, the resulting views may agree with one another while expressing an edit that is weak or insufficiently faithful to the instruction.

Our key observation is that keyframe selection has a substantial impact on the final editing quality. This suggests that view-guided editing requires more than simply designating a set of anchors: the source–target roles should be assigned according to edit-specific reliability and maintained throughout both multi-view editing and subsequent 3D optimization. A view that provides clearer geometric, semantic, and edit-related evidence should provide guidance for a less informative view.

Based on this observation, we propose \textbf{View Matters}. Keyframe Importance Estimation (KIE) determines which views provide reliable evidence for the requested edit. Keyframe-Guided Editing (KGE) transfers editing signals from these keyframes to non-keyframes without noisy reverse influence. Importance-Aware Optimization (IAO) preserves the same reliability preference during 3DGS optimization, preventing reliable supervision from being diluted when edited views are projected back to the 3D representation. Shared keyframe guidance and local propagation maintain cross-view coherence throughout this process.

Our contributions are threefold:
\begin{itemize}
\item We formulate text-driven 3DGS editing as an unequal-view-reliability problem and propose KIE to identify reliable editing anchors using geometric, semantic, and edit-related cues.

\item We introduce KGE to propagate reliable editing signals from keyframes to non-keyframes without reverse contamination and IAO to preserve the same reliability prior during 3DGS optimization.

\item Across 23 scene-prompt pairs, View Matters achieves the best average CLIP text--image and directional similarities among the evaluated methods with a four-minute editing time. Ablations verify the contribution of all three modules, while adjacent-view analysis indicates that improved fidelity does not sacrifice cross-view coherence.
\end{itemize}

\section{Related Work}

\subsection{2D Image Editing}

Pretrained diffusion models provide strong semantic priors for text-guided image editing. P2P~\cite{hertz2022prompt} controls edits through cross-attention, while ControlNet~\cite{zhang2023adding} introduces additional structural conditions. InstructPix2Pix (IP2P)~\cite{brooks2023instructpix2pix} performs instruction-based image editing without requiring a user-provided target image. We adopt a frozen IP2P editor as the editing backbone and reorganize its attention for reliability-guided multi-view interaction.

\subsection{Text-driven 3D Editing}

Due to the scarcity of large-scale paired 3D editing data, most text-driven 3D editing methods transfer the priors of pretrained 2D diffusion models to 3D. Related 3D generation methods, such as DreamFusion~\cite{poole2022dreamfusion} and ConsistentDreamer~\cite{csahin2025consistentdreamer}, synthesize new 3D content from text or a single image. Although they do not edit existing reconstructions, they demonstrate the effectiveness of diffusion guidance for 3D optimization. For editing existing scenes, NeRF-based approaches either optimize the underlying representation using score distillation or related diffusion objectives, as in Instruct 3D-to-3D~\cite{kamata2023instruct}, DreamEditor~\cite{zhuang2023dreameditor}, and Progressive3D~\cite{cheng2023progressive3d}, or follow a render-edit-optimize pipeline, as exemplified by Instruct-NeRF2NeRF~\cite{haque2023instruct} and related follow-ups~\cite{liu2024genn2n,wang2022clip,wang2023nerf}. These methods establish the feasibility of transferring 2D editing priors to 3D, but NeRF optimization is often slow, while independently edited views may provide inconsistent supervision.

With the emergence of 3DGS, text-driven 3D editing methods increasingly adopt explicit Gaussian representations for more efficient rendering and optimization. GaussianEditor~\cite{wang2024gaussianeditor} improves editing efficiency and controllability by leveraging the explicit structure of Gaussians. More recent methods, including GaussCtrl~\cite{wu2024gaussctrl}, DGE~\cite{chen2024dge}, EditSplat~\cite{lee2025editsplat}, and D$^2$Gaussian~\cite{sheng2025d2gaussian}, further improve multi-view consistency or editing control through depth conditioning and latent alignment, epipolar-guided feature propagation, multi-view fusion, and discretized view modeling, respectively. These methods primarily improve how editing information is aligned or propagated across views. View Matters instead explicitly estimates which views provide reliable evidence for the requested edit and preserves this view-reliability hierarchy throughout both multi-view editing and 3D optimization.

\subsection{View Selection and Multi-View Interaction}
View selection has been used to reduce redundancy and improve multi-view reconstruction \cite{hornung2008image}. Neural rendering methods quantify useful observations through uncertainty reduction, Fisher information, or reconstruction-aware criteria \cite{pan2022activenerf,jiang2024fisherrf,xiao2024nerf}. Their objective is to acquire or retain views that best reconstruct a scene. Text-driven editing poses a different question: a useful view must expose scene geometry and carry distinctive content and evidence relevant to a language instruction.

Multi-frame diffusion methods commonly reuse attention features from reference frames to stabilize generated content \cite{ceylan2023pix2video}. We adopt this mechanism but organize the interaction asymmetrically: a small set of keyframes interacts globally and then guides non-keyframes together with local context. This one-way guidance prevents less informative views from weakening reliable keyframe signals.

\begin{figure*}[t]
  \centering
  \includegraphics[width=\textwidth]{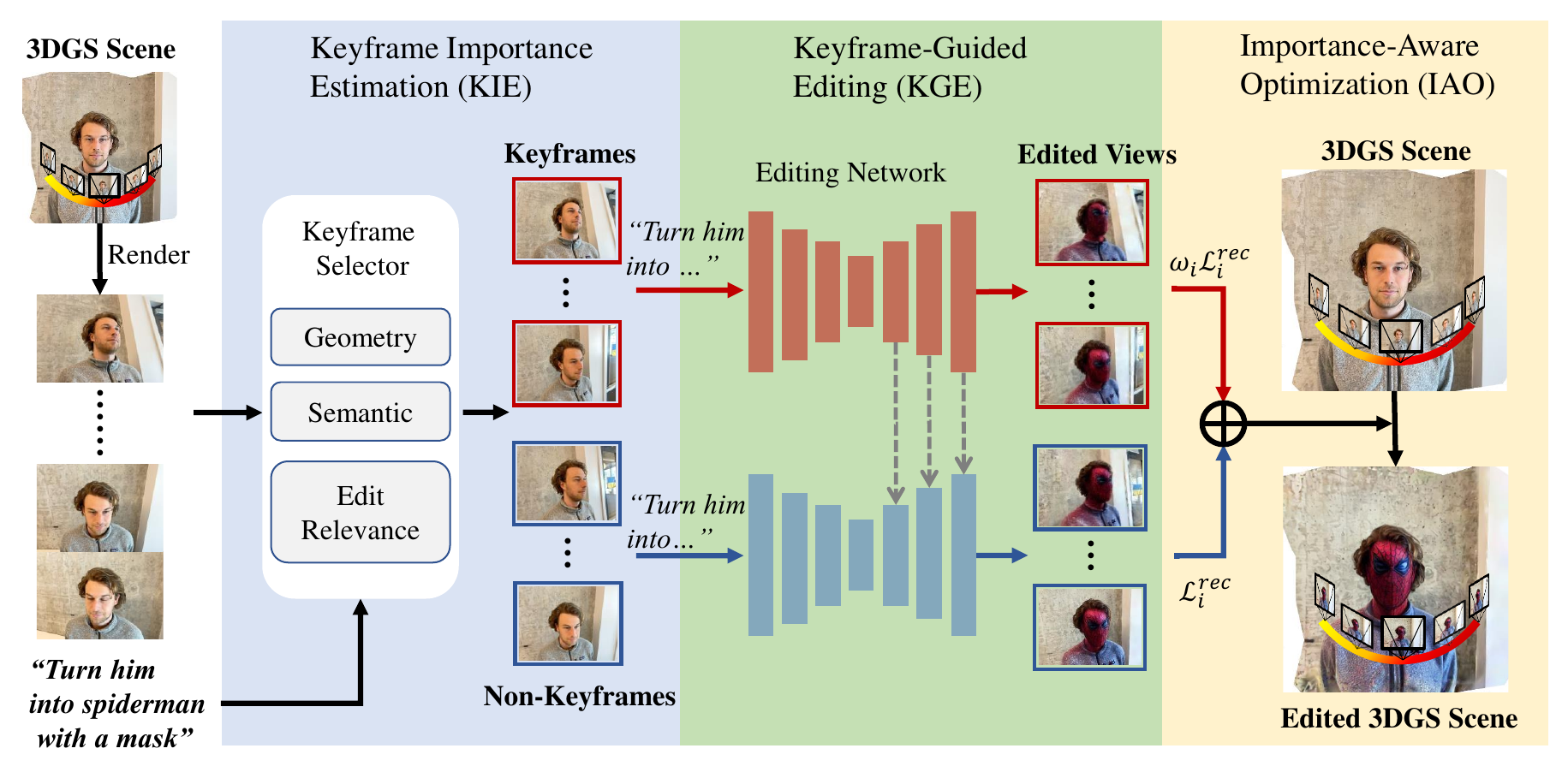}
  \caption{Overview of View Matters. Given a 3DGS scene and an editing instruction, KIE estimates view reliability and selects representative keyframes. KGE lets these keyframes establish a shared context and asymmetrically guide less-reliable non-keyframes without reverse interference. IAO then preserves the reliability hierarchy by assigning stronger supervision to keyframes during 3DGS optimization. The resulting scene follows the requested edit faithfully while maintaining cross-view coherence.}
  \label{fig:pipeline}
\end{figure*}

\section{Method}

\subsection{Problem Formulation}
Given a 3D Gaussian scene
$G=\{g_m\}_{m=1}^{M}$ and camera views
$V=\{v_i\}_{i=1}^{N}$, the renderer produces
$\mathbf{x}_i=\mathcal{R}(G,v_i)$. Given an editing instruction
$\mathbf{P}$, a pretrained image editor produces an edited observation
$\tilde{\mathbf{x}}_i=\mathcal{E}(\mathbf{x}_i,\mathbf{P})$.
Our method partitions the view set into keyframes $V^k$ and
non-keyframes $V^n=V\setminus V^k$, which provide different levels
of supervision during editing and 3D optimization.

In addition to RGB images, the renderer also provides view-dependent geometric cues, such as depth and visibility, under each view. These cues support view-importance estimation.

Besides the edited image itself, we denote by
\begin{equation}
\mathbf{F}_i = \Phi(\mathbf{x}_i, \mathbf{P})
\end{equation}
the intermediate feature representation extracted from the editor, where $\Phi(\cdot)$ represents the feature extraction process.

\subsection{Overview}
As shown in Figure~\ref{fig:pipeline}, given a 3D Gaussian scene and a text prompt, our goal is to produce an edited 3D scene that is both semantically faithful to the instruction and consistent across views. Instead of treating all rendered views equally, we perform the editing process under the guidance of a small set of representative keyframes. Our framework consists of three stages: 1) \textbf{Keyframe Importance Estimation (KIE)}, designed to score rendered views and determine representative keyframes; 2) \textbf{Keyframe-Guided Editing (KGE)}, developed to propagate editing signals from reliability-selected keyframes to non-keyframes through directed attention; and 3) \textbf{Importance-Aware Optimization (IAO)}, introduced to prioritize representative keyframes via loss reweighting while retaining the original reconstruction objective.

\subsection{Keyframe Importance Estimation (KIE)}
Keyframes serve as the main source of editing guidance, a natural question is which rendered views should be selected as keyframes. We address this question by estimating view importance from three complementary perspectives: geometric visibility, semantic distinctiveness~\cite{radford2021learning}, and edit relevance~\cite{fu2023guiding}. The normalized scores are then fused into a final importance score, based on which representative keyframes are selected.

\paragraph{Geometric Visibility.}
Because a 3DGS representation is reconstructed from multi-view observations, a view that observes more Gaussians covers a larger portion of the reconstructed geometry and can provide broader geometric evidence for editing. We therefore use the ratio of visible Gaussians as a geometry-based cue of view importance. Let $M$ denote the total number of Gaussians, and let $\mathbb{1}(\cdot)$ be the visibility indicator provided by the 3DGS renderer~\cite{kerbl20233d}. The geometry visibility score of view $v_i$ is defined as
\begin{equation}
s_i^{\mathrm{geo}}
=
\frac{1}{M}
\sum_{m=1}^{M}
\mathbb{1}(g_m \text{ is visible in } v_i),
\end{equation}
which corresponds to the ratio of visible Gaussians under view $v_i$.

\paragraph{Semantic Distinctiveness.}
Semantically distinctive views are more likely to be informative keyframes, as they capture unusual content that cannot be readily inferred from other views. We then measure how distinctive each view is in the semantic feature space. Let $\phi_{\mathrm{img}}(\cdot)$ denote the CLIP image encoder~\cite{radford2021learning}. We extract normalized image features for all candidate views and compute their mean feature:
\begin{equation}
\mathbf{f}_i
=
\frac{\phi_{\mathrm{img}}(\mathbf{x}_i)}
{\|\phi_{\mathrm{img}}(\mathbf{x}_i)\|_2},
\qquad
\bar{\mathbf{f}}
=
\frac{1}{N}\sum_{i=1}^{N}\mathbf{f}_i.
\end{equation}
The semantic distinctiveness score is then defined as the Euclidean distance between the feature of view $v_i$ and the mean feature:
\begin{equation}
s_i^{\mathrm{sem}}
=
\|\mathbf{f}_i-\bar{\mathbf{f}}\|_2.
\end{equation}
A larger value indicates that the view is more distinctive relative to the overall view distribution.

\begin{figure*}[t]
  \centering
  \includegraphics[width=\textwidth]{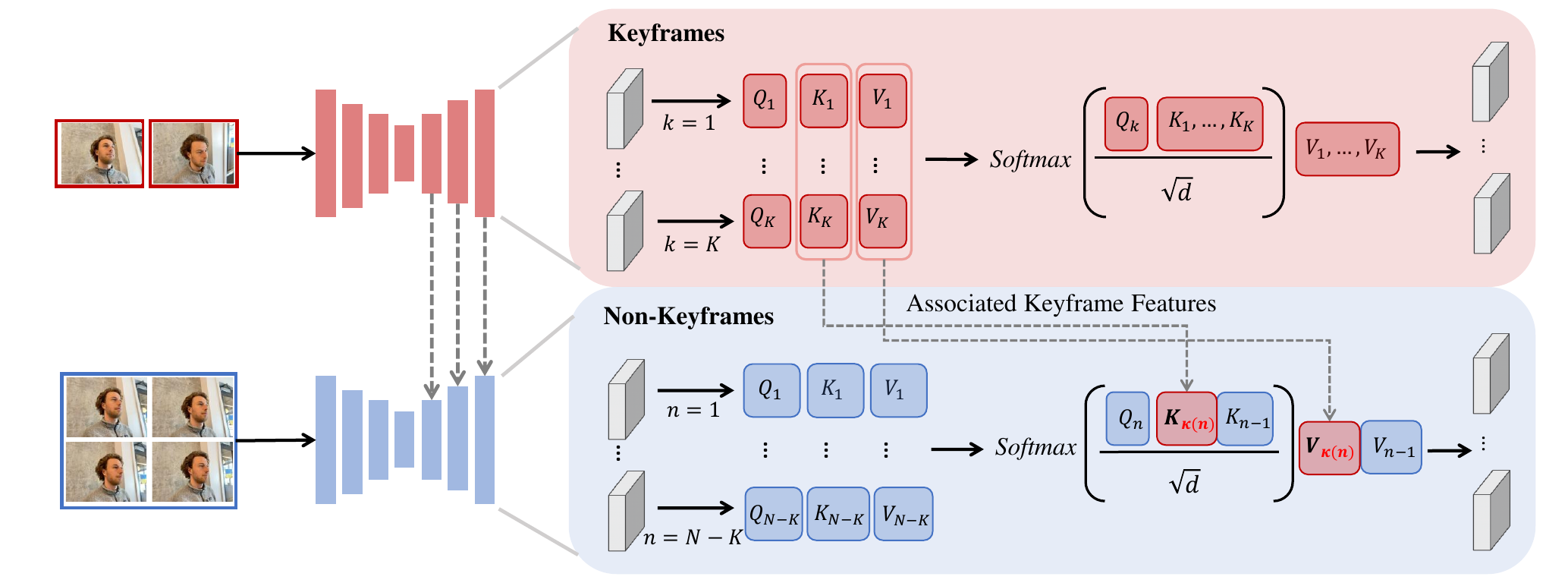}
  \caption{Illustration of the proposed keyframe-guided attention mechanism. Our method modifies the attention computation inside a frozen IP2P denoising network without introducing additional trainable parameters. Keyframes first perform global interaction to establish a shared editing context. Each non-keyframe then attends to its associated keyframe together with the immediately preceding frame, yielding an asymmetrical propagation scheme that uses keyframes as the dominant guidance source while maintaining local continuity across views.}
  \label{fig:attention}
\end{figure*}

\paragraph{Edit Relevance.}
We also consider views that are more closely aligned with the editing instruction to be more suitable keyframes, as they provide more direct guidance for the intended transformation. To estimate how relevant a view is to the target editing instruction, we compute the similarity between the image feature and the text feature in the CLIP embedding space. Let $\phi_{\mathrm{txt}}(\cdot)$ denote the CLIP text encoder~\cite{radford2021learning}, and let
\begin{equation}
\mathbf{t}
=
\frac{\phi_{\mathrm{txt}}(\mathbf{P})}
{\|\phi_{\mathrm{txt}}(\mathbf{P})\|_2}
\end{equation}
be the normalized text feature of the editing prompt $\mathbf{P}$. The edit relevance score is defined as
\begin{equation}
s_i^{\mathrm{edit}}
=
\mathbf{f}_i^{\top}\mathbf{t}.
\end{equation}

\paragraph{Score Fusion.}
The three cues are complementary, as a view weak in one aspect may remain informative in others. We therefore fuse them for robust keyframe selection. Each cue is independently min-max normalized across all candidate views, with $\epsilon=10^{-8}$ for numerical stability. Denoting the normalized scores by $\hat{s}_i^{(\cdot)}$, we compute the importance of view $v_i$ as
\begin{equation}
s_i =
w_{\mathrm{geo}}\hat{s}_i^{\mathrm{geo}}
+w_{\mathrm{sem}}\hat{s}_i^{\mathrm{sem}}
+w_{\mathrm{edit}}\hat{s}_i^{\mathrm{edit}}.
\end{equation}
We use $(w_{\mathrm{geo}},w_{\mathrm{sem}},w_{\mathrm{edit}})
=(0.6,0.2,0.2)$ by default.

\paragraph{Group-wise Keyframe Selection.}
To ensure broad spatial coverage, views are ordered along the rendering trajectory and partitioned into $K$ contiguous groups:
\begin{equation}
\{\mathcal{G}_k\}_{k=1}^{K},
\qquad
\bigcup_{k=1}^{K}\mathcal{G}_k = V.
\end{equation}
For each group $\mathcal{G}_k$, we select the view with the highest importance score as the representative keyframe. The resulting keyframe subset is defined as
\begin{equation}
V^{k}=\left\{\,v^{*}_{k}\;\middle|\; v^{*}_{k}=\arg\max_{i:\,v_i\in \mathcal{G}_k} s_i,\; k=1,\dots,K \right\},
\end{equation}
and the remaining views form the non-keyframe subset
\begin{equation}
V^{n}
=
V\setminus V^{k}.
\end{equation}

\subsection{Keyframe-Guided Editing (KGE)}
Given the selected keyframes, we perform multi-view editing by modifying the attention computation in a pretrained IP2P model. We do not introduce additional trainable parameters. Instead, the IP2P network is kept frozen, and keyframe-guided multi-view interaction is imposed directly within its attention layers during the denoising process. Figure~\ref{fig:attention} illustrates the proposed keyframe-guided attention mechanism.

\paragraph{Global Interaction among Keyframes.}
We first let the selected keyframes interact globally to establish a coherent editing context. For each keyframe $v_{k}\in V^k$, we use its query, key, and value features from the frozen IP2P attention layer, denoted by
$\mathbf{Q}_{k}, \mathbf{K}_{k}, \mathbf{V}_{k}.$
We then aggregate information across all keyframes by
\begin{equation}
\bar{\mathbf{F}}_{k}
=
\mathrm{Attn}\!\left(\mathbf{Q}_{k}, \mathbf{K}^{k}, \mathbf{V}^{k}\right),
\end{equation}
where
\begin{equation}
\mathbf{K}^{k}=[\mathbf{K}_{1},\ldots,\mathbf{K}_{K}], \qquad
\mathbf{V}^{k}=[\mathbf{V}_{1},\ldots,\mathbf{V}_{K}],
\end{equation}
and
\begin{equation}
\label{Equ:attention}
\mathrm{Attn}(\mathbf{Q},\mathbf{K},\mathbf{V})
=
\mathrm{softmax}\!\left(\frac{\mathbf{Q}\mathbf{K}^{\top}}{\sqrt{d}}\right)\mathbf{V}.
\end{equation}
This step builds a globally consistent semantic context among representative keyframes.

\paragraph{Asymmetrical Propagation to Non-keyframes.}
For each non-keyframe $v_n\in V^n$, we modify the IP2P attention computation in the decoder layers~\cite{ceylan2023pix2video} so that it receives editing guidance asymmetrically from its associated keyframe, rather than interacting symmetrically with all views. Specifically, $\kappa(n)\in V^k$ denotes the keyframe selected from the group containing $v_n$. The group keyframe provides reliable editing guidance, whereas the previous frame offers stronger local correspondence because adjacent views typically share greater visual overlap. We therefore use $v_{n-1}$, when available, as an auxiliary guidance source to preserve local cross-view continuity. The propagated feature of $v_n$ is computed as
\begin{equation}
\hat{\mathbf{F}}_n
=
\mathrm{Attn}\!\left(
\mathbf{Q}_n,
\mathbf{K}_n^{\mathrm{mem}},
\mathbf{V}_n^{\mathrm{mem}}
\right),
\end{equation}
where the memory bank is defined as
\begin{equation}
\begin{aligned}
\mathbf{K}^{\mathrm{mem}}_n &=
\begin{cases}
[\mathbf{K}_{\kappa(n)}, \mathbf{K}_{n-1}], & n>1,\\
[\mathbf{K}_{\kappa(n)}], & n=1,
\end{cases}\\
\mathbf{V}^{\mathrm{mem}}_n &=
\begin{cases}
[\mathbf{V}_{\kappa(n)}, \mathbf{V}_{n-1}], & n>1,\\
[\mathbf{V}_{\kappa(n)}], & n=1.
\end{cases}
\end{aligned}
\end{equation}

This design is inherently asymmetric: the associated keyframe provides the primary editing guidance, while the previous frame supplies additional local context for smoother propagation. For the first frame, the propagation naturally degenerates to keyframe-only guidance. Since the modification is applied directly to frozen IP2P attention layers, the editing process remains training-free.

\paragraph{Edited Multi-view Observations.}
After the modified denoising process, the frozen IP2P model directly produces the edited result for each view. We denote the final edited observation of view $v_i$ by
\begin{equation}
\mathbf{y}_i = \mathcal{E}_{\mathrm{IP2P}}(\mathbf{x}_i,\mathbf{P}),
\end{equation}
where $\mathcal{E}_{\mathrm{IP2P}}(\cdot)$ denotes the frozen IP2P editor with our keyframe-guided attention modification. For keyframes, the editing result is obtained under global keyframe interaction. For non-keyframes, the result is further guided by asymmetrical propagation from the associated keyframe and local neighboring views.

\subsection{Importance-Aware Optimization (IAO)}
KIE identifies reliable keyframes, and KGE uses them to guide the remaining views. However, treating all edited views equally during 3DGS optimization would discard this reliability ordering and allow less reliable edits to weaken keyframe supervision. We therefore introduce IAO to preserve keyframe priority throughout scene optimization.

After obtaining edited multi-view observations, we optimize the underlying 3D Gaussian scene using a reconstruction-based objective, while assigning higher optimization priority to representative keyframes through a lightweight loss reweighting strategy. Instead of redesigning the entire optimization pipeline, we keep the original training objective and only increase the contribution of keyframe views during optimization.

In practice, training is performed with batch size 1. Let $v_i$ denote the currently sampled view, and let
\begin{equation}
\mathbf{x}_i^{\,r} = \mathcal{R}(G, v_i)
\end{equation}
be the rendered image of the current 3D Gaussian scene. Let $\mathbf{y}_i$ denote the corresponding edited supervision image cached for view $v_i$. We first compute a reconstruction loss between the current rendering and the edited target:
\begin{equation}
\mathcal{L}_i^{\mathrm{rec}}
=
\lambda_{l1}\mathcal{L}_{l1}(\mathbf{x}_i^{\,r}, \mathbf{y}_i)
+
\lambda_{p}\mathcal{L}_{\mathrm{perc}}(\mathbf{x}_i^{\,r}, \mathbf{y}_i),
\end{equation}
where $\mathcal{L}_{l1}$ is the pixel-wise $\ell_1$ loss~\cite{isola2017image} and $\mathcal{L}_{\mathrm{perc}}$ is a perceptual loss~\cite{zhang2018unreasonable}.

To emphasize the role of representative keyframes, we apply a keyframe-aware weighting factor to the reconstruction loss of the current training step:
\begin{equation}
\tilde{\mathcal{L}}_i = \omega_i \mathcal{L}_i^{\mathrm{rec}},
\end{equation}
where
\begin{equation}
\omega_i=
\begin{cases}
\lambda_k,(\lambda_k>1) & v_i \in V^k, \\
1, & v_i \in V^n,
\end{cases}
\end{equation}
and $\lambda_k$ is the keyframe loss weight. In this way, the optimization process remains identical to the original reconstruction-based pipeline except that keyframe views receive stronger supervision.

\begin{figure*}[t]
  \centering
  \includegraphics[page=1, width=\textwidth]{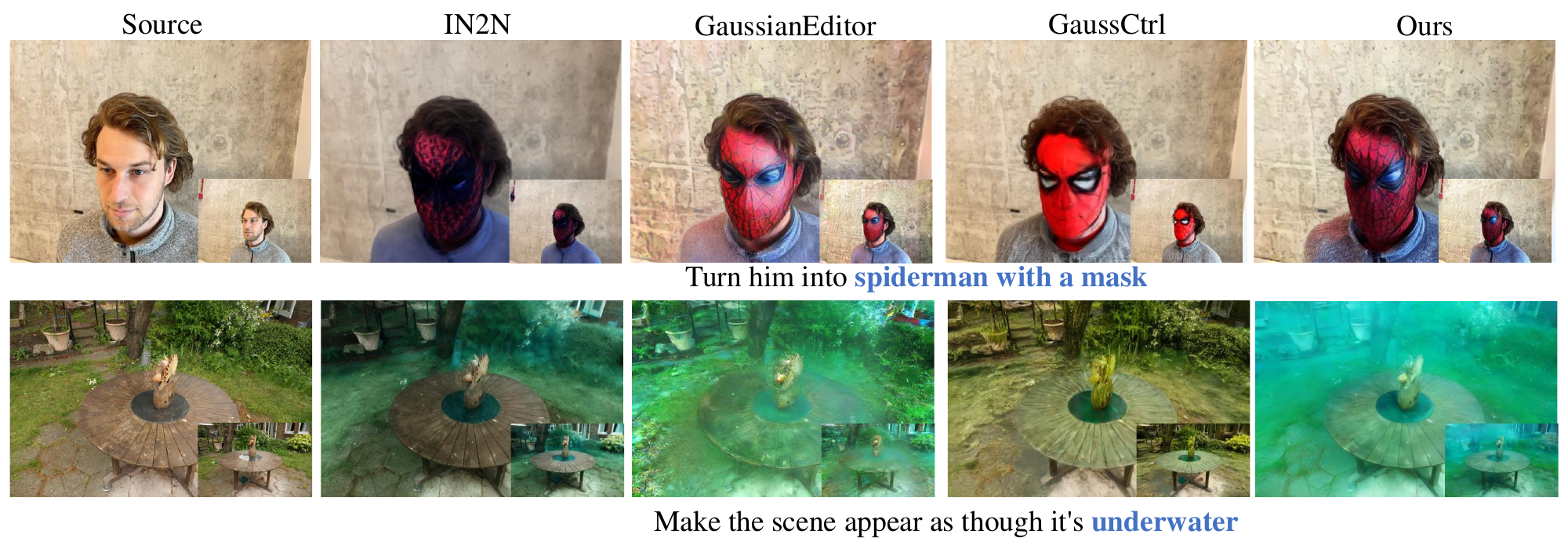}
  \caption{The two examples span complementary editing regimes: a localized edit of a human subject (top) and a global transformation of an outdoor scene (bottom). Our method more faithfully follows both instructions while preserving editing-irrelevant content and the underlying scene structure. Additional comparisons across scenes and edit instructions are provided in Appendix~\ref{sec:supp_results}.}
  \label{fig:qualitative}
\end{figure*}

\section{Experiments}
In this section, we first introduce the experimental setup and compare our method with representative approaches. We then directly examine how keyframe selection affects editing quality, analyze the contributions and sensitivity of the proposed components, and finally verify that the fidelity-oriented design preserves cross-view coherence.

\subsection{Experimental Setup}
\subsubsection{Datasets and Metrics.} Following prior works, we select subsets from Mip-NeRF 360~\cite{barron2022mip} and IN2N~\cite{haque2023instruct} for editing evaluation. We adopt CLIP text-image similarity (T-I)~\cite{radford2021learning} and CLIP directional similarity (Direction)~\cite{gal2022stylegan} to measure the alignment between edited results and text instructions. In addition, we use adjacent-view CLIP temporal scores~\cite{haque2023instruct} to evaluate cross-view consistency, which is largely overlooked in previous methods. The evaluation contains 23 unique scene-prompt pairs covering object identity and appearance, material transformation, global environment and style, and localized editing.

\subsubsection{Baselines.} We compare our method with representative NeRF-based approaches, including IN2N~\cite{haque2023instruct} and VICA~\cite{dong2023vica}, as well as representative 3DGS-based approaches, including GaussianEditor~\cite{wang2024gaussianeditor}, GaussCtrl~\cite{wu2024gaussctrl}, DGE~\cite{chen2024dge}, and EditSplat~\cite{lee2025editsplat}. These 3DGS baselines cover iterative dataset updating, depth- and attention-based alignment, epipolar-constrained key-view propagation, and multi-view fusion, respectively.

\subsubsection{Implementation Details.} Our method is implemented in PyTorch and can be optimized on a single 48G RTX 4090 GPU. For both NeRF-based and 3DGS-based baselines, the initial 3D representations are reconstructed from the same multi-view images under identical training settings. For fair comparison, all methods use the same input scenes, text prompts, and initialization whenever applicable. Additional details on hyperparameters, scene selection, and prompts are provided in Appendix~\ref{sec:supp_impl}.

\subsection{Quantitative Analysis}
\begin{table}[t]
\centering
\small
\setlength{\tabcolsep}{3pt}
\renewcommand{\arraystretch}{1.08}

\begin{tabular}{@{}lccc@{}}
\toprule
Method & T--I $\uparrow$ & Direction $\uparrow$ & Time $\downarrow$ \\
\midrule
IN2N           & 0.2599              & 0.1767              & 28 min \\
VICA           & 0.2439              & 0.1565              & 25 min \\
GaussianEditor & \underline{0.2769}  & \underline{0.2227}  & 8 min  \\
GaussCtrl      & 0.2589              & 0.1572              & 10 min \\
EditSplat      & 0.2719              & 0.2152              & 7 min  \\
DGE            & 0.2753              & 0.2179              & \underline{5 min}  \\
Ours           & \textbf{0.2822}     & \textbf{0.2564}     & \textbf{4 min} \\
\bottomrule
\end{tabular}
\caption{Quantitative comparison with representative text-driven 3D editing methods.
Best results are in bold, and second-best results are underlined.}
\label{tab:quantitative_comparison}
\end{table}

Table~\ref{tab:quantitative_comparison} shows that View Matters achieves the best average T-I (0.2822) and Direction (0.2564) among the evaluated methods. The directional score improves by 0.0337 over GaussianEditor and 0.0412 over EditSplat, indicating more accurate instruction-following changes. Our method also completes editing in four minutes, faster than the evaluated baselines. These results show the overall benefit of organizing editing around reliable views.

\subsection{Qualitative Analysis}
Figure~\ref{fig:qualitative} presents two representative cases selected to span complementary regimes in both content type and spatial extent: localized editing of a human subject and global appearance editing of an outdoor scene. The former tests whether a method can apply the requested semantic change without unnecessarily modifying the surrounding content, whereas the latter requires the edit to propagate coherently across the scene while preserving its underlying geometry and layout.

In these examples, IN2N distorts the source content, GaussianEditor modifies editing-irrelevant regions, and GaussCtrl only partially realizes the requested semantics. In contrast, our method handles both regimes by localizing the subject-level edit and coherently propagating the scene-level transformation, achieving a better balance between instruction fidelity and content preservation. Additional comparisons are provided in Appendix~\ref{sec:supp_results}.

\begin{figure}[t]
  \centering
  \includegraphics[width=\linewidth]{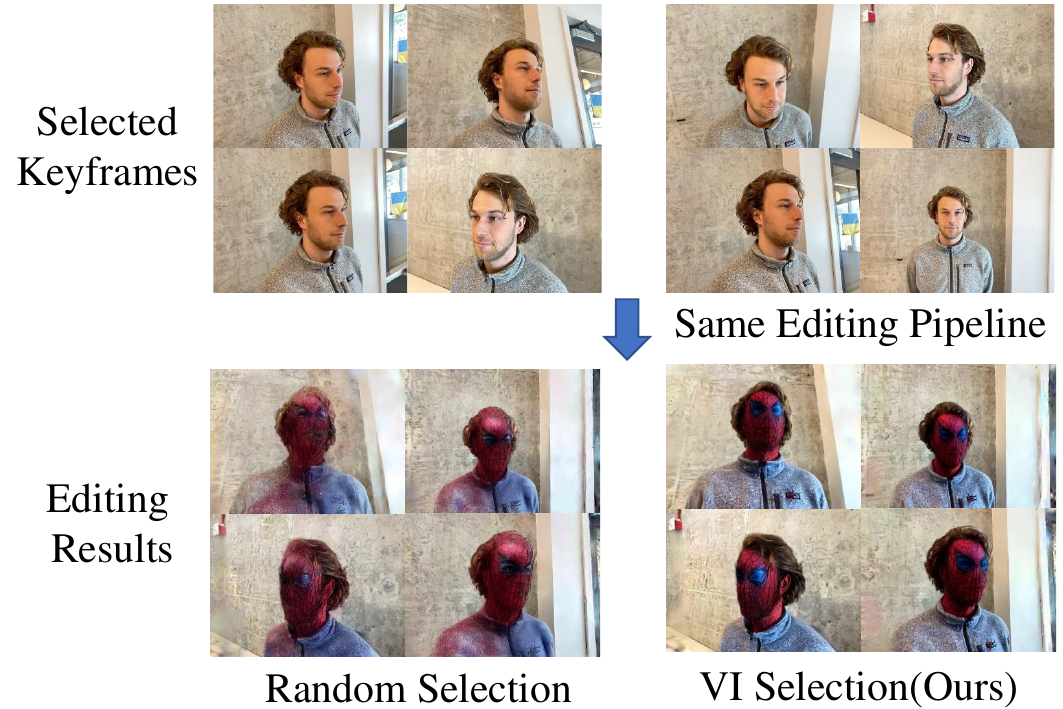}
  \caption{With the same candidate views and keyframe budget, random and importance-based selection yield different keyframes (top) and editing results (bottom). Our selection provides more reliable guidance and achieves a more faithful target transformation, highlighting the importance of view selection in multi-view editing.}
  \label{fig:view}
\end{figure}

\begin{figure}[t]
  \centering
  \includegraphics[width=\linewidth]{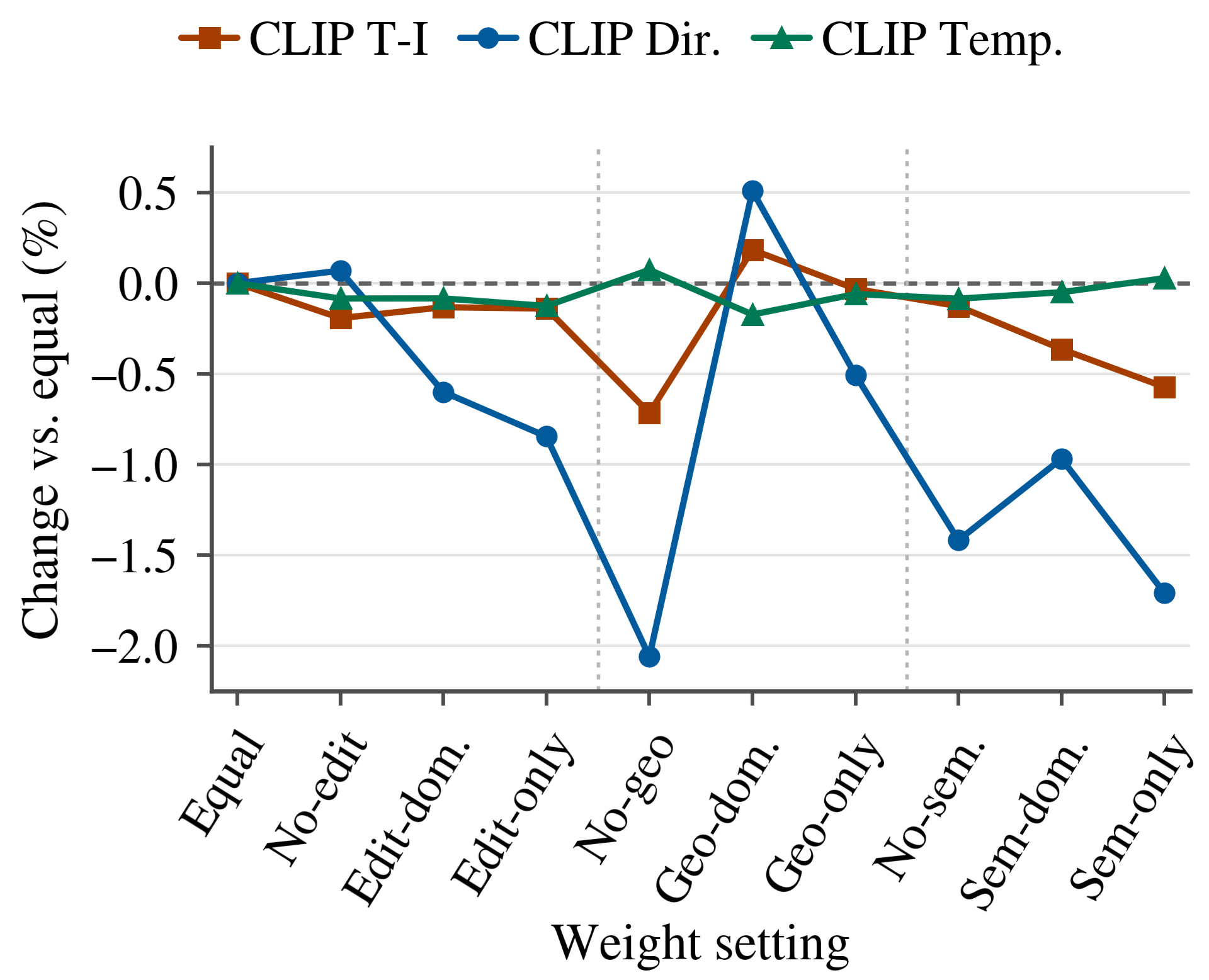}
  \caption{Sensitivity of view-importance weights. The three cues are complementary, with Geo-dom.\ $(0.6,0.2,0.2)$ adopted as the default for its best overall fidelity.}
  \label{fig:sen_score}
\end{figure}

\subsection{Analysis of View Importance}
To directly test our central claim that views contribute unequally to multi-view editing, we use a controlled sparse-view setting in which the effect of keyframe selection can be more clearly isolated. We compare importance-based selection with random selection while keeping the candidate views, keyframe budget, editing procedure, and 3DGS optimization identical. Thus, the only experimental variable is which views are selected as keyframes.

Figure~\ref{fig:view} shows both the selected keyframes and their corresponding editing results. Random sampling may select views that provide less reliable geometric, semantic, or editability cues, thereby weakening the editing signal propagated to the remaining views. In contrast, our method identifies more informative keyframes and uses them to guide the other views, producing a more faithful target transformation while maintaining coherent appearance across viewpoints.

The results show that editing quality depends not only on how many views are used but also on which views provide the guidance. This result supports our central argument that views should not be treated equally in multi-view editing.

\subsection{Ablation and Sensitivity Analysis}
Following this controlled comparison, we progressively add KGE, KIE, and IAO to the per-view IP2P baseline. With KGE alone, the first view in each group serves as the keyframe. Table~\ref{tab:ablation} shows that KGE improves performance through keyframe propagation, KIE provides more informative guidance, and IAO performs best by preserving keyframe priority during 3DGS optimization. These gains confirm their complementary roles. Qualitative ablations are provided in Appendix~\ref{sec:supp_ablation}.

Figure~\ref{fig:sen_score} analyzes the geometric visibility, semantic distinctiveness, and edit-relevance cues in KIE. Here, $X$-dom assigns 0.6 to cue $X$ and 0.2 to the others; $X$-only retains only $X$; and No-$X$ removes $X$ while equally weighting the other two cues. Removing the geometric cue causes the largest degradation in editing fidelity. This cue is derived directly from the underlying 3DGS and captures the intrinsic view-dependent observability of the scene, providing a reliable basis for identifying informative views. Accordingly, Geo-dom, with weights $(0.6,0.2,0.2)$, achieves the highest TI and Direction while largely preserving adjacent-view consistency. This result supports our central design of explicitly selecting views according to their 3D editing evidence. Nevertheless, geometry alone underperforms Geo-dom: removing semantics weakens the intended edit direction, while removing edit relevance reduces semantic alignment and consistency. These results demonstrate that the three cues are complementary. We therefore adopt Geo-dom as the default setting, prioritizing 3D geometric evidence while retaining semantic and instruction-related cues for keyframe selection.

\begin{table}[t]
\centering
\small
\setlength{\tabcolsep}{3.5pt}
\begin{tabular}{ccc|cc}
\toprule
KIE & KGE & IAO & T-I $\uparrow$ & Direction $\uparrow$ \\
\midrule
 &  &  & 0.2702 & 0.2229 \\
 & \checkmark &  & 0.2754 & 0.2464 \\
\checkmark & \checkmark &  & 0.2761 & 0.2486 \\
\checkmark & \checkmark & \checkmark & \textbf{0.2822} & \textbf{0.2564} \\
\bottomrule
\end{tabular}
\caption{Ablation study on View Matters. \checkmark\ denotes the inclusion of each module. The full model consistently achieves the best results.}
\label{tab:ablation}
\end{table}

\begin{figure}[t]
  \centering
  \includegraphics[width=\linewidth]{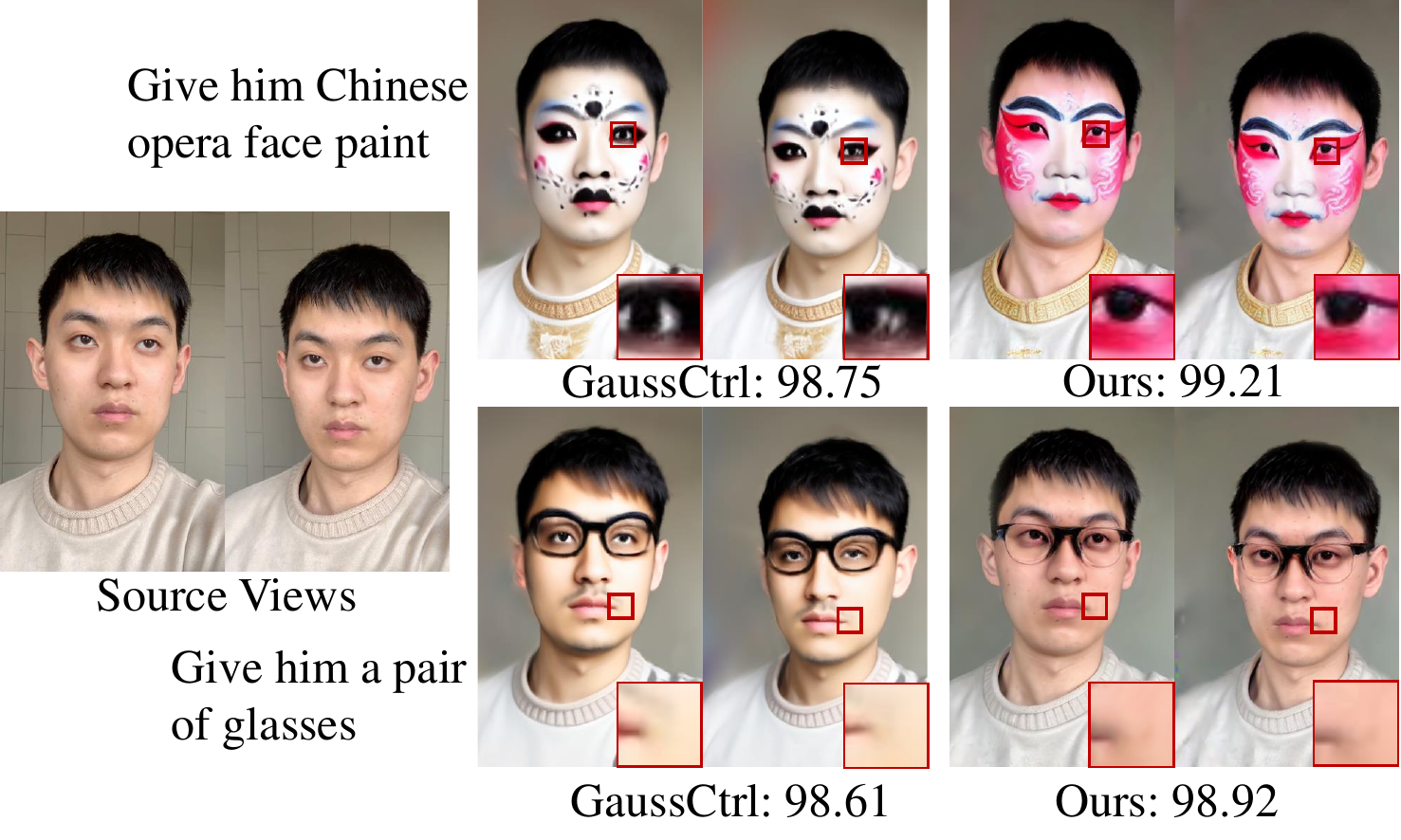}
  \caption{Representative adjacent-view consistency analysis on two ordered rendering sequences.}
  \label{fig:temp_score}
\end{figure}

\subsection{Cross-View Consistency Analysis}
Although View Matters introduces no separate consistency-specific module or loss, its reliability-guided design does not sacrifice cross-view coherence. Instead, shared keyframe guidance in KGE establishes a common editing context, while local propagation maintains continuity between adjacent views. Figure~\ref{fig:temp_score} presents two representative ordered-view case studies rather than an aggregate consistency benchmark. Quantitatively, our method achieves higher CLIP temporal scores in both examples, reflecting stronger consistency across rendered views. Qualitatively, this improvement is evident in the highlighted regions, where our method produces more coherent local details and more stable visual patterns across viewpoints. In contrast, GaussCtrl suffers from greater inter-view variation, leading to inconsistent local appearance and reduced structural coherence.

\section{Conclusion}
View Matters organizes text-driven 3D Gaussian editing around unequal view reliability. KIE identifies reliable editing anchors, KGE propagates their information asymmetrically without noisy reverse influence, and IAO preserves the same priority during 3D optimization. The resulting framework improves instruction alignment and editing efficiency among the evaluated methods, while representative adjacent-view analyses indicate that the fidelity-oriented design maintains cross-view coherence.

\bibliographystyle{plainnat}
\bibliography{references}

@article{kerbl20233d,
  title={3d gaussian splatting for real-time radiance field rendering.},
  author={Kerbl, Bernhard and Kopanas, Georgios and Leimk{\"u}hler, Thomas and Drettakis, George},
  journal={ACM Trans. Graph.},
  volume={42},
  number={4},
  pages={139--1},
  year={2023}
}

@article{hertz2022prompt,
  title={Prompt-to-prompt image editing with cross attention control},
  author={Hertz, Amir and Mokady, Ron and Tenenbaum, Jay and Aberman, Kfir and Pritch, Yael and Cohen-Or, Daniel},
  journal={arXiv preprint arXiv:2208.01626},
  year={2022}
}

@inproceedings{zhang2023adding,
  title={Adding conditional control to text-to-image diffusion models},
  author={Zhang, Lvmin and Rao, Anyi and Agrawala, Maneesh},
  booktitle={Proceedings of the IEEE/CVF international conference on computer vision},
  pages={3836--3847},
  year={2023}
}

@inproceedings{brooks2023instructpix2pix,
  title={Instructpix2pix: Learning to follow image editing instructions},
  author={Brooks, Tim and Holynski, Aleksander and Efros, Alexei A},
  booktitle={Proceedings of the IEEE/CVF conference on computer vision and pattern recognition},
  pages={18392--18402},
  year={2023}
}

@article{poole2022dreamfusion,
  title={Dreamfusion: Text-to-3d using 2d diffusion},
  author={Poole, Ben and Jain, Ajay and Barron, Jonathan T and Mildenhall, Ben},
  journal={arXiv preprint arXiv:2209.14988},
  year={2022}
}

@article{kamata2023instruct,
  title={Instruct 3d-to-3d: Text instruction guided 3d-to-3d conversion},
  author={Kamata, Hiromichi and Sakuma, Yuiko and Hayakawa, Akio and Ishii, Masato and Narihira, Takuya},
  journal={arXiv preprint arXiv:2303.15780},
  year={2023}
}

@inproceedings{zhuang2023dreameditor,
  title={Dreameditor: Text-driven 3d scene editing with neural fields},
  author={Zhuang, Jingyu and Wang, Chen and Lin, Liang and Liu, Lingjie and Li, Guanbin},
  booktitle={SIGGRAPH Asia 2023 conference papers},
  pages={1--10},
  year={2023}
}

@article{cheng2023progressive3d,
  title={Progressive3d: Progressively local editing for text-to-3d content creation with complex semantic prompts},
  author={Cheng, Xinhua and Yang, Tianyu and Wang, Jianan and Li, Yu and Zhang, Lei and Zhang, Jian and Yuan, Li},
  journal={arXiv preprint arXiv:2310.11784},
  year={2023}
}

@inproceedings{haque2023instruct,
  title={Instruct-nerf2nerf: Editing 3d scenes with instructions},
  author={Haque, Ayaan and Tancik, Matthew and Efros, Alexei A and Holynski, Aleksander and Kanazawa, Angjoo},
  booktitle={Proceedings of the IEEE/CVF international conference on computer vision},
  pages={19740--19750},
  year={2023}
}

@inproceedings{wang2024gaussianeditor,
  title={Gaussianeditor: Editing 3d gaussians delicately with text instructions},
  author={Wang, Junjie and Fang, Jiemin and Zhang, Xiaopeng and Xie, Lingxi and Tian, Qi},
  booktitle={Proceedings of the IEEE/CVF conference on computer vision and pattern recognition},
  pages={20902--20911},
  year={2024}
}

@inproceedings{wu2024gaussctrl,
  title={Gaussctrl: Multi-view consistent text-driven 3d gaussian splatting editing},
  author={Wu, Jing and Bian, Jia-Wang and Li, Xinghui and Wang, Guangrun and Reid, Ian and Torr, Philip and Prisacariu, Victor Adrian},
  booktitle={European conference on computer vision},
  pages={55--71},
  year={2024},
  organization={Springer}
}

@inproceedings{chen2024dge,
  title={Dge: Direct gaussian 3d editing by consistent multi-view editing},
  author={Chen, Minghao and Laina, Iro and Vedaldi, Andrea},
  booktitle={European conference on computer vision},
  pages={74--92},
  year={2024},
  organization={Springer}
}

@inproceedings{lee2025editsplat,
  title={Editsplat: Multi-view fusion and attention-guided optimization for view-consistent 3d scene editing with 3d gaussian splatting},
  author={Lee, Dong In and Park, Hyeongcheol and Seo, Jiyoung and Park, Eunbyung and Park, Hyunje and Baek, Ha Dam and Shin, Sangheon and Kim, Sangmin and Kim, Sangpil},
  booktitle={Proceedings of the Computer Vision and Pattern Recognition Conference},
  pages={11135--11145},
  year={2025}
}

@inproceedings{sheng2025d2gaussian,
  title={D2Gaussian: Dynamic Control with Discretized 3D View Modeling for Text-Driven 3D Gaussian Splatting Editing},
  author={Sheng, Yefei and Wang, Jie and Tao, Ming and Bao, Bing-Kun},
  booktitle={Proceedings of the 33rd ACM International Conference on Multimedia},
  pages={9296--9305},
  year={2025}
}

@inproceedings{hornung2008image,
  title={Image selection for improved multi-view stereo},
  author={Hornung, Alexander and Zeng, Boyi and Kobbelt, Leif},
  booktitle={2008 IEEE Conference on Computer Vision and Pattern Recognition},
  pages={1--8},
  year={2008},
  organization={IEEE}
}

@inproceedings{pan2022activenerf,
  title={Activenerf: Learning where to see with uncertainty estimation},
  author={Pan, Xuran and Lai, Zihang and Song, Shiji and Huang, Gao},
  booktitle={European Conference on Computer Vision},
  pages={230--246},
  year={2022},
  organization={Springer}
}

@inproceedings{jiang2024fisherrf,
  title={Fisherrf: Active view selection and mapping with radiance fields using fisher information},
  author={Jiang, Wen and Lei, Boshu and Daniilidis, Kostas},
  booktitle={European Conference on Computer Vision},
  pages={422--440},
  year={2024},
  organization={Springer}
}

@inproceedings{xiao2024nerf,
  title={Nerf director: Revisiting view selection in neural volume rendering},
  author={Xiao, Wenhui and Cruz, Rodrigo Santa and Ahmedt-Aristizabal, David and Salvado, Olivier and Fookes, Clinton and Lebrat, Leo},
  booktitle={Proceedings of the IEEE/CVF Conference on Computer Vision and Pattern Recognition},
  pages={20742--20751},
  year={2024}
}

@inproceedings{radford2021learning,
  title={Learning transferable visual models from natural language supervision},
  author={Radford, Alec and Kim, Jong Wook and Hallacy, Chris and Ramesh, Aditya and Goh, Gabriel and Agarwal, Sandhini and Sastry, Girish and Askell, Amanda and Mishkin, Pamela and Clark, Jack and others},
  booktitle={International conference on machine learning},
  pages={8748--8763},
  year={2021},
  organization={PmLR}
}

@inproceedings{barron2022mip,
  title={Mip-nerf 360: Unbounded anti-aliased neural radiance fields},
  author={Barron, Jonathan T and Mildenhall, Ben and Verbin, Dor and Srinivasan, Pratul P and Hedman, Peter},
  booktitle={Proceedings of the IEEE/CVF conference on computer vision and pattern recognition},
  pages={5470--5479},
  year={2022}
}

@article{gal2022stylegan,
  title={Stylegan-nada: Clip-guided domain adaptation of image generators},
  author={Gal, Rinon and Patashnik, Or and Maron, Haggai and Bermano, Amit H and Chechik, Gal and Cohen-Or, Daniel},
  journal={ACM Transactions on Graphics (TOG)},
  volume={41},
  number={4},
  pages={1--13},
  year={2022},
  publisher={ACM New York, NY, USA}
}

@article{dong2023vica,
  title={Vica-nerf: View-consistency-aware 3d editing of neural radiance fields},
  author={Dong, Jiahua and Wang, Yu-Xiong},
  journal={Advances in Neural Information Processing Systems},
  volume={36},
  pages={61466--61477},
  year={2023}
}

@article{csahin2025consistentdreamer,
  title={ConsistentDreamer: View-consistent meshes through balanced multi-view Gaussian optimization},
  author={{\c{S}}ahin, Onat and Altillawi, Mohammad and Eskandar, George and Carbone, Carlos and Liu, Ziyuan},
  journal={Pattern Recognition Letters},
  volume={190},
  pages={118--125},
  year={2025},
  publisher={Elsevier}
}

@inproceedings{liu2024genn2n,
  title={Genn2n: Generative nerf2nerf translation},
  author={Liu, Xiangyue and Xue, Han and Luo, Kunming and Tan, Ping and Yi, Li},
  booktitle={Proceedings of the IEEE/CVF Conference on Computer Vision and Pattern Recognition},
  pages={5105--5114},
  year={2024}
}

@inproceedings{wang2022clip,
  title={Clip-nerf: Text-and-image driven manipulation of neural radiance fields},
  author={Wang, Can and Chai, Menglei and He, Mingming and Chen, Dongdong and Liao, Jing},
  booktitle={Proceedings of the IEEE/CVF conference on computer vision and pattern recognition},
  pages={3835--3844},
  year={2022}
}

@article{wang2023nerf,
  title={Nerf-art: Text-driven neural radiance fields stylization},
  author={Wang, Can and Jiang, Ruixiang and Chai, Menglei and He, Mingming and Chen, Dongdong and Liao, Jing},
  journal={IEEE Transactions on Visualization and Computer Graphics},
  volume={30},
  number={8},
  pages={4983--4996},
  year={2023},
  publisher={IEEE}
}

@article{fu2023guiding,
  title={Guiding instruction-based image editing via multimodal large language models},
  author={Fu, Tsu-Jui and Hu, Wenze and Du, Xianzhi and Wang, William Yang and Yang, Yinfei and Gan, Zhe},
  journal={arXiv preprint arXiv:2309.17102},
  year={2023}
}

@inproceedings{zhang2018unreasonable,
  title={The unreasonable effectiveness of deep features as a perceptual metric},
  author={Zhang, Richard and Isola, Phillip and Efros, Alexei A and Shechtman, Eli and Wang, Oliver},
  booktitle={Proceedings of the IEEE conference on computer vision and pattern recognition},
  pages={586--595},
  year={2018}
}

@inproceedings{isola2017image,
  title={Image-to-image translation with conditional adversarial networks},
  author={Isola, Phillip and Zhu, Jun-Yan and Zhou, Tinghui and Efros, Alexei A},
  booktitle={Proceedings of the IEEE conference on computer vision and pattern recognition},
  pages={1125--1134},
  year={2017}
}

@inproceedings{ceylan2023pix2video,
  title={Pix2video: Video editing using image diffusion},
  author={Ceylan, Duygu and Huang, Chun-Hao P and Mitra, Niloy J},
  booktitle={Proceedings of the IEEE/CVF international conference on computer vision},
  pages={23206--23217},
  year={2023}
}

@article{tao2025c3editor,
  title={C3Editor: Achieving Controllable Consistency in 2D Model for 3D Editing},
  author={Tao, Zeng and Ding, Zheng and Chen, Zeyuan and Zhang, Xiang and Li, Leizhi and Tu, Zhuowen},
  journal={arXiv preprint arXiv:2510.04539},
  year={2025}
}

@article{bengtson2025threeDconsistent,
  title={{3D-Consistent Multi-View Editing by Diffusion Guidance}},
  author={Bengtson, Josef and Nilsson, David and Lee, Dong In and Kahl, Fredrik},
  journal={arXiv preprint arXiv:2511.22228},
  year={2025}
}
\clearpage
\onecolumn
\appendix
\section*{Appendix Overview}
\label{sec:supp_overview}

The appendices provide additional implementation details, ablation experiments, qualitative comparisons, and discussions that complement the main paper. They are organized as follows:

\begin{itemize}
    \item Appendix~\ref{sec:supp_impl} provides additional implementation details and hyperparameter settings.

    \item Appendix~\ref{sec:supp_ablation} presents additional ablation studies on the number of keyframes and the keyframe-loss weight in IAO, followed by qualitative ablation results.

    \item Appendix~\ref{sec:supp_results} provides comparisons with recent methods without public implementations, evaluates generalization across human subjects, non-human objects, and complete scenes, and presents additional qualitative results.

    \item Appendix~\ref{sec:supp_limitations} discusses the limitations of the current framework and possible directions for future work.
\end{itemize}

\section{Additional Implementation Details}
\label{sec:supp_impl}
We follow the parameter settings used in previous methods~\cite{haque2023instruct,wang2024gaussianeditor,chen2024dge}.
Unless otherwise specified, for most scenes, the guidance scales are set to 7.5 for the textual condition and 1.5 for the image condition.
For each scene, we use 20 edited views, including 4 keyframes and 16 non-keyframes, and adopt a batch size of 5 during optimization.
The selected views are updated every 500 iterations.
The weight of the keyframe loss is set to 1.6.
For evaluation, as shown in Table~\ref{tab:detailed_dataset} we select 8 scenes and construct 23 scene-prompt pairs.
For each prompt, we report the average score across all rendered views.

\section{Additional Ablation Experiments}
\label{sec:supp_ablation}
In this section, we conduct additional ablation studies to justify the chosen hyperparameter settings and to further verify the effectiveness of the three components in the keyframe selection module.
\subsection{Impact of the Number of Keyframes}
We set the total number of edited views to 20~\cite{wang2024gaussianeditor,chen2024dge}, following previous work. With the total number of views fixed, varying the interval between selected keyframes is equivalent to varying the total number of keyframes. We therefore evaluate four settings, corresponding to 2, 4, 5, and 10 keyframes.
\begin{appendixtable}
\centering
\small
\setlength{\tabcolsep}{5pt}
\renewcommand{\arraystretch}{1.1}
\begin{tabular}{c c c c c c}
\hline
\textbf{Number of keyframes} & \textbf{T-I $\uparrow$} & \textbf{Dir $\uparrow$} & \textbf{Training Time (s)} & \textbf{Max Alloc. (GB)} & \textbf{Max Res. (GB)} \\
\hline
2  & 0.2763   & 0.2781   & 263.46 & 17.78 & 25.38 \\
4  & \textbf{0.2793} & \textbf{0.2840} & 266.04 & 17.78 & 25.37 \\
5  & 0.2786   & 0.2821   & 268.34 & 17.78 & 25.39 \\
10 & 0.2780   & 0.2819   & 324.58 & 24.54 & 37.25 \\
\hline
\end{tabular}
\caption{\textbf{Effectiveness and resource consumption under different keyframe settings.}}
\label{tab:resource_editviews}
\end{appendixtable}

As shown in Tab.~\ref{tab:resource_editviews}, increasing the number of keyframes from 2 to 4 improves both T-I and Dir, while the training time and memory consumption remain almost unchanged. This indicates that introducing a moderate number of keyframes can provide stronger supervision without noticeable additional computational overhead. When the number of keyframes is further increased to 5, the resource cost is still similar, but the performance shows a slight drop compared with the 4-view setting. Increasing the number of keyframes to 10 leads to a clear increase in both training time and GPU memory usage, while failing to bring further quantitative gains. These results suggest that too few keyframes provide insufficient supervision, whereas too many views introduce redundant information and unnecessary computational burden. Therefore, we adopt 4 keyframes as the default setting, since it achieves the best trade-off between editing performance and efficiency.

\clearpage
\subsection{Effect of the Keyframe-Loss Weight in IAO}
Figure~\ref{fig:weight_IAO} analyzes the sensitivity to the keyframe-loss weight $\lambda_k$. The coarse sweep reveals a clear trade-off: increasing $\lambda_k$ generally improves Text--Image Similarity but gradually reduces Directional Similarity, suggesting that excessively strong keyframe supervision may favor semantic alignment at the expense of the intended edit direction. The fine-grained sweep further shows that $\lambda_k=1.6$ provides the best balance. Both metrics improve over $\lambda_k=1$, whereas larger weights yield only marginal gains in Text--Image Similarity while decreasing Directional Similarity. We therefore set $\lambda_k=1.6$ as the default.

\begin{appendixfigure}
  \centering
  \includegraphics[width=0.46\linewidth]{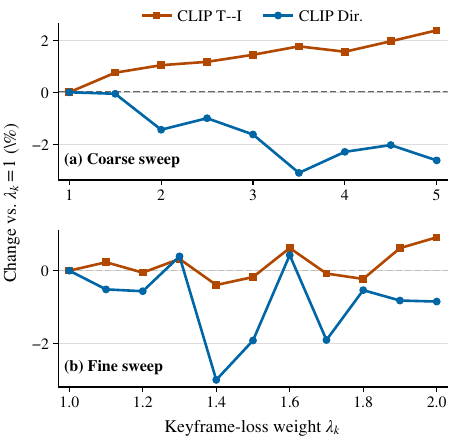}
  \caption{Sensitivity to the keyframe-loss weight $\lambda_k$.
    Relative changes in CLIP T--I and CLIP Directional Similarity are
    reported against $\lambda_k=1$ for the coarse and fine-grained sweeps.}
  \label{fig:weight_IAO}
\end{appendixfigure}
\subsection{Qualitative Ablation Results}
As shown in Fig.~\ref{fig:qua_ablation}, Keyframe-Guided Editing (KGE) introduces richer target textures, suppresses undesired texture changes on the clothes, and mitigates the inconsistency around the neck region. Further incorporating Keyframe Importance Estimation enhances the details of the eye-mask region and removes part of the residual original eye contour. After additionally introducing Importance-Aware Optimization (IAO), the method fully suppresses the interference of the original eye contours and better preserves the appearance of non-edited regions, yielding the most reliable and consistent editing results. This improvement indicates that only when the 3D optimization process explicitly accounts for the different importance of views can the model achieve the best editing quality, which also validates our core insight that view importance matters in 3D Gaussian editing.

\begin{appendixfigure}
  \centering
  \includegraphics[width=0.74\linewidth]{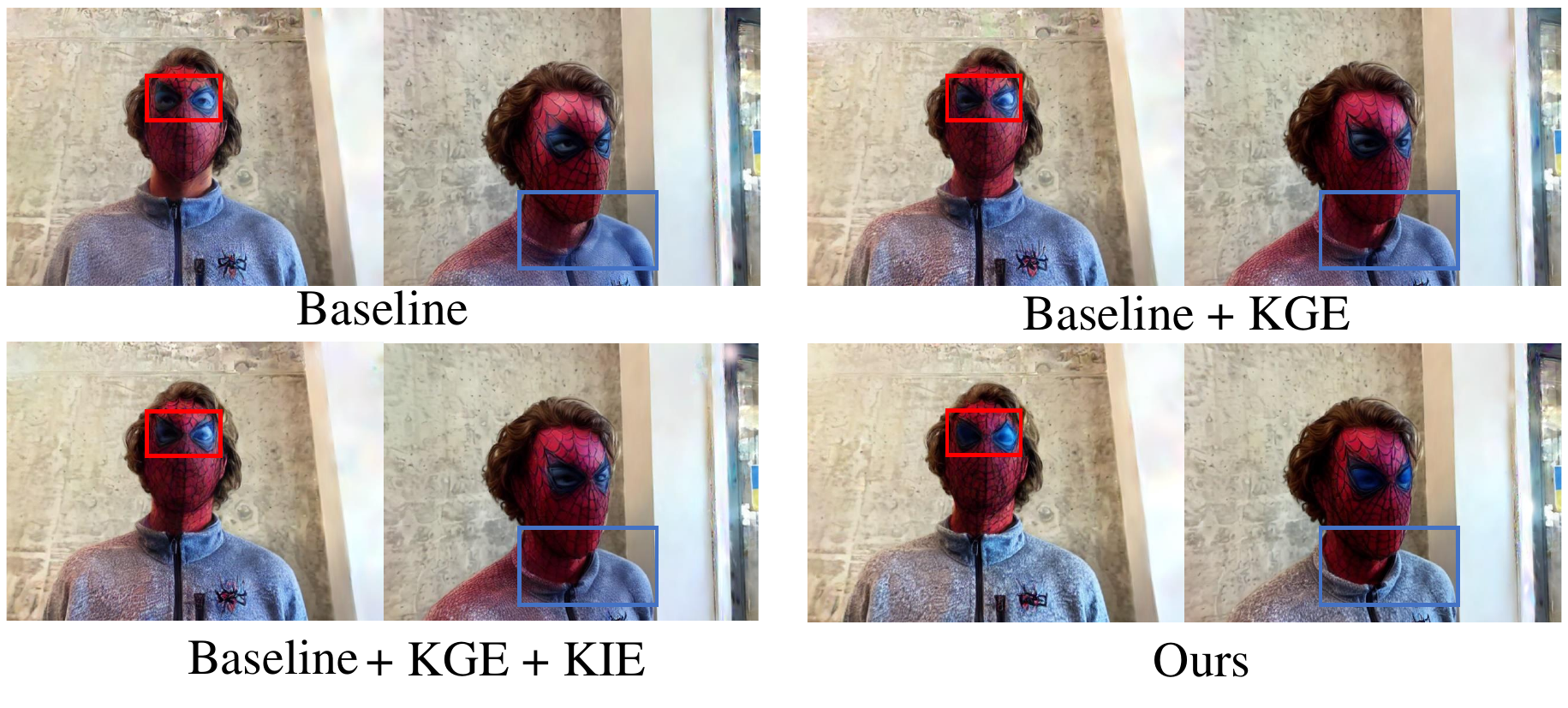}
  \caption{Qualitative ablation results. The progressive visual improvements illustrate the complementary effects of KGE, KIE, and IAO. Red and blue boxes highlight different local regions for comparison.}
  \label{fig:qua_ablation}
\end{appendixfigure}
\clearpage
\section{Additional Experiment Results}
\label{sec:supp_results}
\subsection{Comparison with Recent Methods without Public Implementations}
Several recent text-driven 3D editing methods~\cite{tao2025c3editor, bengtson2025threeDconsistent} have not released their official implementations, preventing a controlled quantitative comparison under the same experimental setting. To provide a qualitative reference, we apply our method using the edit prompts reported in their papers and compare the resulting visual effects in Fig.~\ref{fig:addbaseline}. Results of the compared methods are taken from their original papers. Although this comparison does not constitute a strictly controlled evaluation, it illustrates how our method handles similar editing instructions while preserving the source content.

\begin{appendixfigure}
    \centering
    \includegraphics[width=0.68\linewidth]{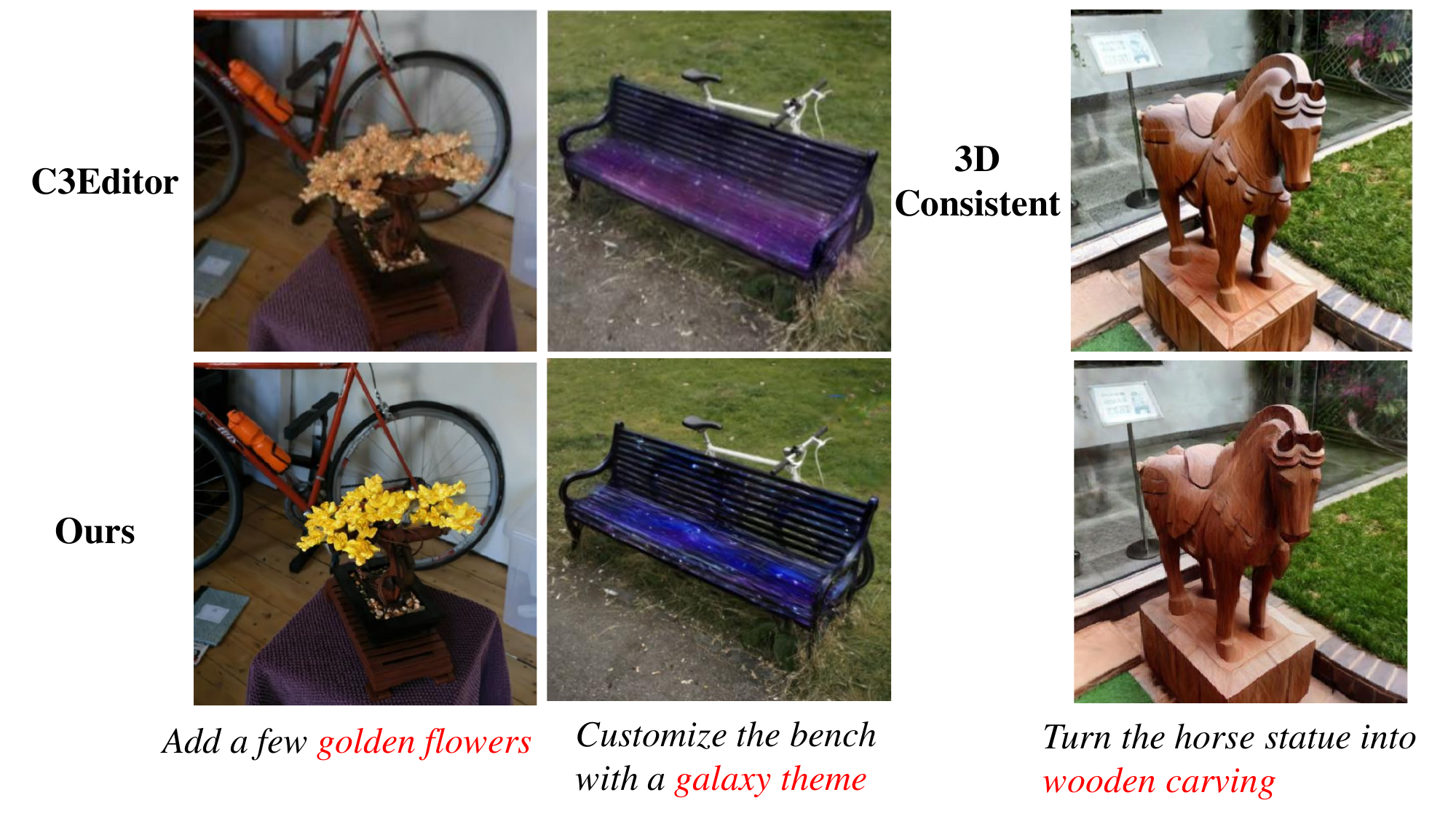}
    \caption{Qualitative comparison with C3Editor and 3D-consistent}
    \label{fig:addbaseline}
\end{appendixfigure}
\subsection{Generalization across Subjects and Edit Types}
Figure~\ref{fig:first} presents results covering human subjects, non-human objects, and complete scenes. These examples include both localized subject edits and global scene transformations, requiring different levels of semantic and structural modification. Our method produces faithful target changes while retaining editing-irrelevant content across these categories, supporting its generalization to diverse subjects and editing scopes.

\begin{appendixfigure}
  \centering
  \includegraphics[width=\textwidth]{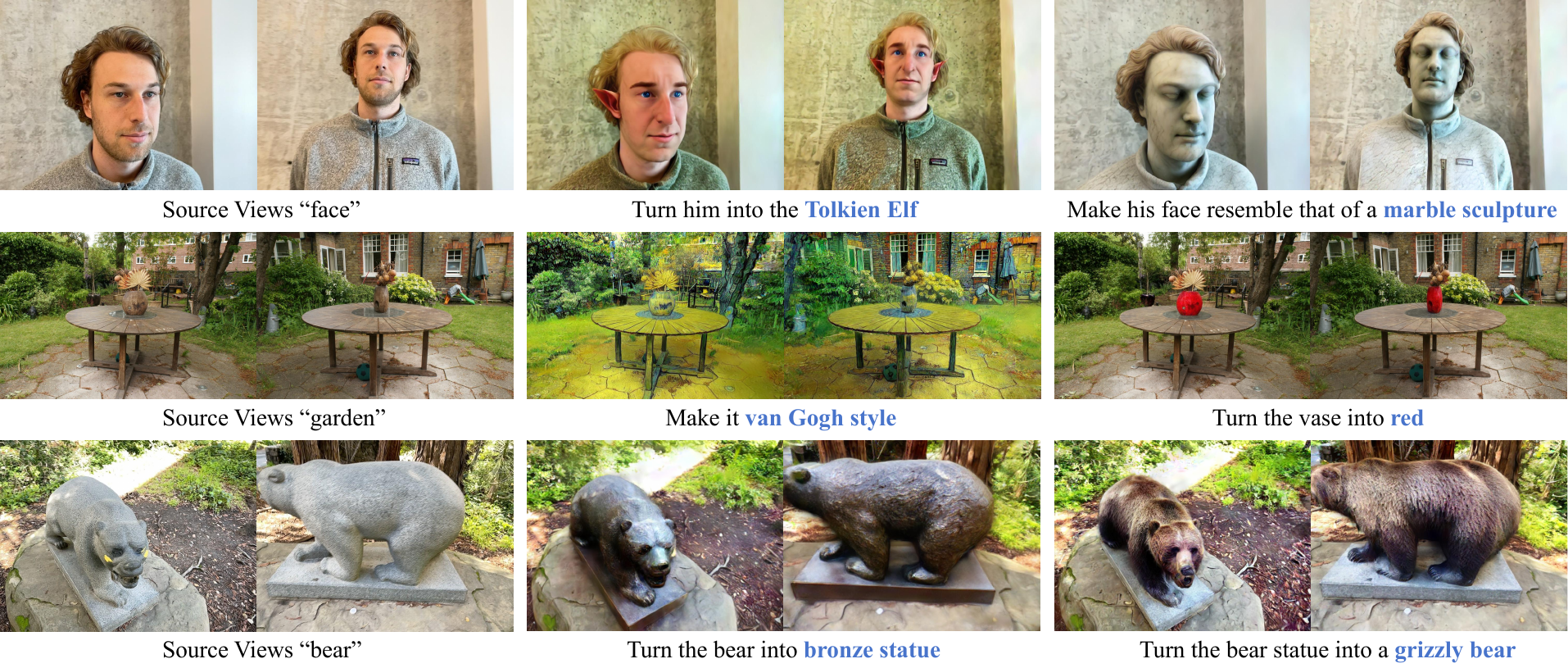}
  \caption{Qualitative results across subjects and edit
scopes. The examples cover human subjects, complete scenes, and
non-human objects, including both localized edits and global
transformations. For each instruction, two rendered viewpoints
show that our method follows the target semantics while preserving
editing-irrelevant content and the underlying scene structure.}
  \label{fig:first}
\end{appendixfigure}
\clearpage
\subsection{Additional Qualitative Results}
Figure~\ref{fig:addition} provides additional results across different scenes and editing instructions. The results further show that our method can express the requested semantic changes while preserving the original scene structure and appearance. Renderings from multiple viewpoints also exhibit coherent edited content, indicating that reliability-guided propagation maintains stable editing effects across views.

\begin{appendixfigure}
  \centering
  \includegraphics[page=1,width=\textwidth]{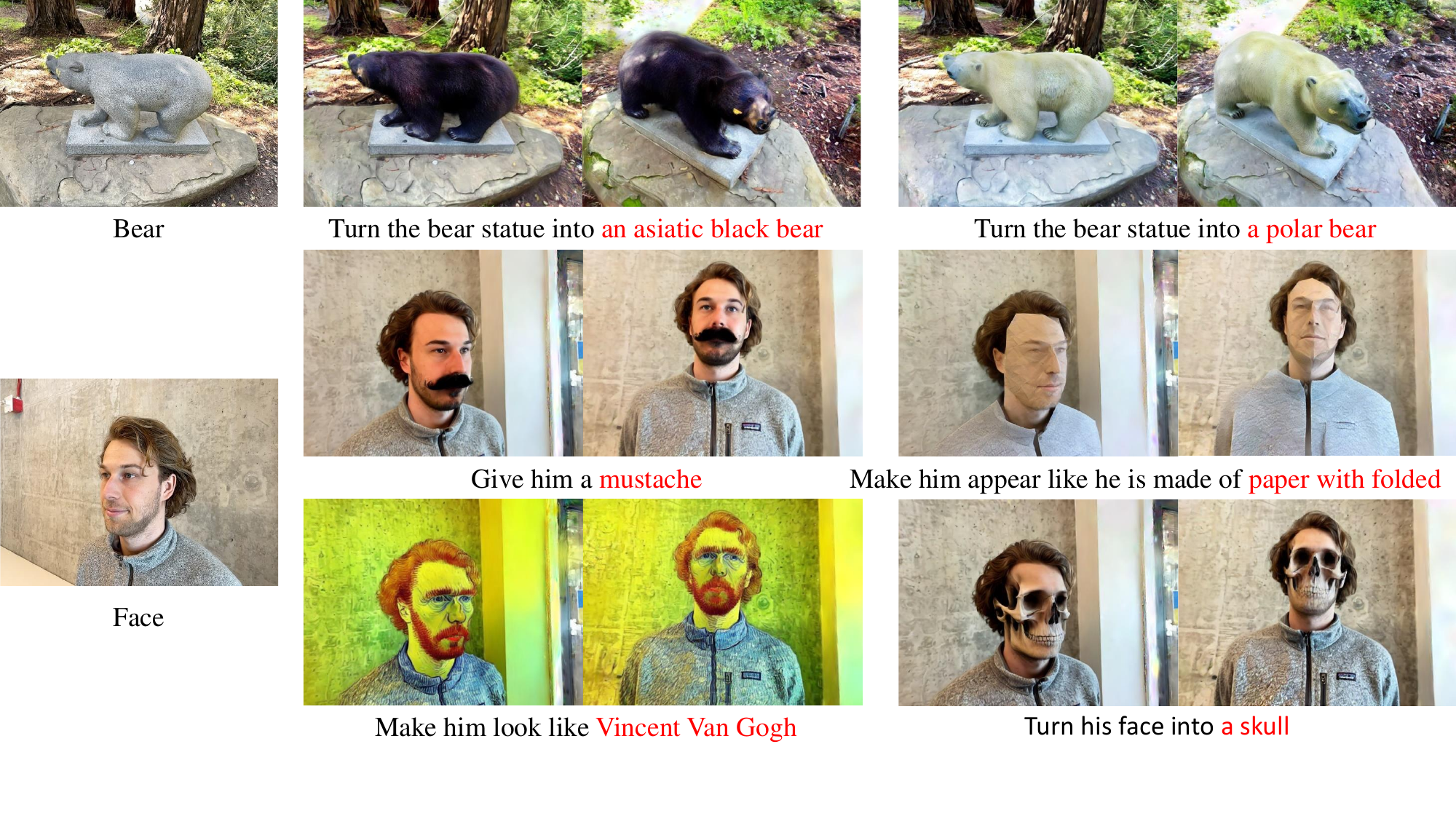}
  \includegraphics[page=2,width=\textwidth]{Figures/supplement/addtional_results.pdf}
  \caption{Additional qualitative results across scenes and editing
instructions. Each group presents the source scene and edited
renderings from multiple viewpoints. Our method handles diverse
identity, material, appearance, style, and color transformations
while maintaining stable edited content across views.}
  \label{fig:addition}
\end{appendixfigure}
\clearpage
\subsection{Evaluation Scenes and Prompts}
\begingroup
\small
\setlength{\tabcolsep}{4pt}
\renewcommand{\arraystretch}{1.02}
\begin{longtable}{@{}>{\raggedright\arraybackslash}p{0.07\textwidth}
>{\raggedright\arraybackslash}p{0.28\textwidth}
>{\raggedright\arraybackslash}p{0.30\textwidth}
>{\raggedright\arraybackslash}p{0.29\textwidth}@{}}
\caption{Detailed evaluation dataset: 23 scene--prompt pairs. The source and target prompts are used for computing CLIP and CLIP directional scores.}\label{tab:detailed_dataset}\\
\hline
\textbf{Scene} & \textbf{Edit Instruction} & \textbf{Target Prompt} & \textbf{Source Prompt} \\\hline
\endfirsthead
\multicolumn{4}{l}{\textit{Table \thetable\ continued}}\\
\hline
\textbf{Scene} & \textbf{Edit Instruction} & \textbf{Target Prompt} & \textbf{Source Prompt} \\\hline
\endhead
\hline
\multicolumn{4}{r}{\textit{Continued on the next page}}\\
\endfoot
\hline
\endlastfoot

Face
& ``Turn him into the Tolkien Elf.''
& ``A Tolkien Elf man with curly hair.''
& ``A man with curly hair in a grey jacket.'' \\

Face
& ``Turn him into an Einstein.''
& ``Einstein with curly hair in a grey jacket.''
& ``A man with curly hair in a grey jacket.'' \\

Face
& ``Turn his face into a skull.''
& ``A skull with curly hair in a grey jacket.''
& ``A man with curly hair in a grey jacket.'' \\

Face
& ``Turn him into spiderman with a mask.''
& ``A spider man with a mask and curly hair.''
& ``A man with curly hair in a grey jacket.'' \\

Person
& ``Turn the man into a clown.''
& ``A clown standing next to a wall wearing a blue T-shirt.''
& ``A man standing next to a wall wearing a blue T-shirt.'' \\

Person
& ``Turn him into a Super Mario.''
& ``A photo of a Super Mario.''
& ``A photo of a person.'' \\

Person
& ``Turn him into a Minecraft character.''
& ``A photo of a person in Minecraft.''
& ``A photo of a person.'' \\

fangzhou
& ``Give him Chinese opera face paint.''
& ``A photo of a face of a man with black hair and Chinese opera face paint.''
& ``A photo of a face of a man with black hair.'' \\

fangzhou
& ``Give him a pair of glasses.''
& ``A photo of a face of a man with black hair wearing a pair of glasses.''
& ``A photo of a face of a man with black hair.'' \\

Bear
& ``Make the color of the bear look like rainbow color.''
& ``A rainbow-colored bear in the forest.''
& ``A stone bear in the forest.'' \\

Bear
& ``Turn the bear into a panda.''
& ``A photo of a panda.''
& ``A photo of a stone bear.'' \\

garden
& ``Make the scene appear as though it's underwater.''
& ``A photo of an outdoor garden in underwater.''
& ``A photo of an outdoor garden.'' \\

garden
& ``Make it Van Gogh style.''
& ``A table in an outdoor garden with Van Gogh style.''
& ``A table in an outdoor garden.'' \\

garden
& ``Make it autumn.''
& ``A photo of an outdoor garden in autumn.''
& ``A photo of an outdoor garden.'' \\

horse statue
& ``Turn the horse statue into a wooden carving.''
& ``A photo of a horse made of wood.''
& ``A photo of a horse statue.'' \\

horse statue
& ``Make the stone horse a zebra.''
& ``A photo of a zebra.''
& ``A photo of a horse statue.'' \\

horse statue
& ``Turn the stone horse into a jade carving.''
& ``A photo of a horse made of jade.''
& ``A photo of a horse statue.'' \\

bonsai
& ``Make the bonsai snowy.''
& ``A photo of a snowy bonsai.''
& ``A photo of a bonsai.'' \\

bonsai
& ``Change the bonsai to look like it's made of paper.''
& ``A photo of a bonsai made of paper.''
& ``A photo of a bonsai.'' \\

bonsai
& ``Add a few golden flowers.''
& ``A photo of a bonsai with golden flowers.''
& ``A photo of a bonsai.'' \\

bicycle
& ``Turn the ground into a Namibian desert.''
& ``A photo of a Namibian desert.''
& ``A photo of a park.'' \\

bicycle
& ``Make the entire scene look as if it's painted in a watercolor style.''
& ``A watercolor style painting of a park.''
& ``A photo of a park.'' \\

bicycle
& ``Customize the bench with a galaxy theme.''
& ``A photo of a bench with a galaxy theme.''
& ``A photo of a bench.'' \\

\end{longtable}
\endgroup

\section{Limitations and Discussion}
\label{sec:supp_limitations}

Our framework has two main limitations. First, it employs a frozen 2D image editor and therefore inherits its failure modes. Edits involving substantial geometric changes or fine-grained local structures may be limited by the underlying editor. Since KGE, KIE, and IAO organize cross-view guidance and 3D supervision without retraining the editing backbone, they could potentially be adapted to stronger attention-based editors in future work.

Second, the benefit of view-importance modeling depends on the variation in editing reliability across candidate views. When densely sampled views provide similarly clear observations, the distinction between keyframes and other views becomes smaller, and the improvement over uniform treatment may be modest. Conversely, if none of the candidate views clearly observes the target region, view selection alone cannot provide a reliable editing anchor. Our controlled sparse-view experiments examine the former reliability differences, while more extreme visibility conditions remain an important direction for future evaluation.

\end{document}